\documentclass[10pt,english,final]{elsarticle}
\usepackage[T1]{fontenc}
\usepackage[latin9]{inputenc}
\usepackage{babel}
\usepackage{array}
\usepackage{booktabs}
\usepackage{url}
\usepackage{multirow}
\usepackage{amsmath}
\usepackage{amssymb}
\usepackage{stackrel}
\usepackage{graphicx}
\usepackage[unicode=true]
 {hyperref}

\makeatletter

\providecommand{\tabularnewline}{\\}

\usepackage{a4wide}
\usepackage{url}
\renewcommand{\citet}{\cite}
\usepackage{times}
\usepackage{algorithmicx}

\makeatother

\begin{document}
\begin{frontmatter}
\journal{arXiv}
\title{CMI: An Online Multi-objective Genetic Autoscaler for Scientific and Engineering Workflows in Cloud Infrastructures with Unreliable Virtual Machines}

\author[itic]{David A. Monge\corref{cor}}
\cortext[cor]{Corresponding author.}
\ead{dmonge@uncu.edu.ar}

\author[itic,conicet]{Elina Pacini}
\ead{epacini@uncu.edu.ar}

\author[isistan]{Cristian Mateos}
\ead{cmateos@conicet.gov.ar}

\author[uma]{Enrique Alba}
\ead{eat@lcc.uma.es}

\author[itic]{Carlos Garc\'ia Garino}
\ead{cgarcia@itu.uncu.edu.ar}

\address[itic]{ITIC, Universidad Nacional de Cuyo. Mendoza, Argentina}


\address[conicet]{Consejo Nacional de Investigaciones Cient\'ificas y T\'ecnicas (CONICET). Argentina.}

\address[isistan]{ISISTAN-UNICEN-CONICET. Tandil, Buenos Aires, Argentina.}

\address[uma]{Departamento de Lenguajes y Ciencias de la Computaci\'on, Universidad de M\'alaga. Spain. }


\begin{abstract}

Cloud Computing is becoming the leading paradigm for executing scientific
and engineering workflows. The large-scale nature of the experiments
they model and their variable workloads make clouds the ideal execution
environment due to prompt and elastic access to huge amounts of computing
resources. Autoscalers are middleware-level software components that
allow scaling up and down the computing platform by acquiring or terminating
virtual machines~(VM) at the time that workflow's tasks are being
scheduled. In this work we propose a novel online multi-objective
autoscaler for workflows denominated Cloud Multi-objective Intelligence~(CMI),
that aims at the minimization of makespan, monetary cost and the potential
impact of errors derived from unreliable VMs. In addition, this problem
is subject to monetary budget constraints. CMI is responsible for
periodically solving the autoscaling problems encountered along the
execution of a workflow. Simulation experiments on four well-known
workflows exhibit that CMI significantly outperforms a state-of-the-art
autoscaler of similar characteristics called Spot Instances Aware
Autoscaling~(SIAA). These results convey a solid base for deepening
in the study of other meta-heuristic methods for autoscaling workflow
applications using cheap but unreliable infrastructures.

\end{abstract}

\begin{keyword}
Cloud Computing, Autoscaling, Scientific workflow, Multi-objective optimization, Evolutionary Algorithm, Unreliable VM Instance
\end{keyword}
\end{frontmatter}

\newcommand{\tasks}[0]{\tau}
\newcommand{\periodtasks}[0]{\tau^{\mathrm{\textit{period}}}}
\newcommand{\deps}[0]{\delta}
\newcommand{\xsca}[0]{X}
\newcommand{\xsch}[0]{X^{\mathrm{\textit{sch}}}}
\newcommand{\vmtypes}[0]{I}
\newcommand{\scavmtypes}[0]{I^{\mathrm{\textit{sca}}}}
\newcommand{\pricemodels}[0]{M}
\newcommand{\instances}[0]{I}

\newcommand{\cmi}[0]{CMI}
\newcommand{\cmiF}[0]{CMI~(full)}
\newcommand{\cmiNs}[0]{CMI~(no-spots)}
\newcommand{\scaf}[0]{SF}
\newcommand{\siaa}[0]{SIAA}

\newcommand{\Mon}[0]{Montage}
\newcommand{\Cyb}[0]{CyberShake}
\newcommand{\Lig}[0]{Inspiral}
\newcommand{\Pan}[0]{Pan-STARRS}

\newcommand{\JavaVersion}[0]{1.9}
\newcommand{\MoeaVersion}[0]{2.12}
\newcommand{\CloudSimVersion}[0]{3.0}
\newcommand{\CommonsMathVersion}[0]{3.3}
\newcommand{\TableauVersion}[0]{2018.3}
\newcommand{\ggplotVersion}[0]{2.2.1}

\newcommand{\vect}[1]{\boldsymbol{#1}}
\newcommand{\vectx}[0]{\vect{x}}
\newcommand{\F}[0]{\vect{f}}

\newcommand{\Xsca}[0]{\vectx^{\mathrm{\textit{sca}}}}
\newcommand{\Xsch}[0]{\vectx^{\mathrm{\textit{sch}}}}

\newcommand{\Xond}[0]{\vectx^{\mathrm{\textit{od}}}}
\newcommand{\Xspt}[0]{\vectx^{\mathrm{\textit{s}}}}
\newcommand{\Xbid}[0]{\vectx^{\mathrm{\textit{b}}}}

\newcommand{\xond}[0]{x^{\mathrm{\textit{od}}}}
\newcommand{\xspt}[0]{x^{\mathrm{\textit{s}}}}
\newcommand{\xbid}[0]{x^{\mathrm{\textit{b}}}}

\newcommand{\objA}[0]{\mathrm{makespan}}
\newcommand{\objB}[0]{\mathrm{cost}}
\newcommand{\objC}[0]{\mathrm{errorsImpact}}

\newcommand{\zideal}[0]{\vect{z}^{\mathrm{\textit{ideal}}}}

\section{\label{sec:Introduction}Introduction}

Today's scientific workflows require large amounts of computational
resources to satisfy the resource-intensive nature of the experiments
they model. A scientific workflow comprises a series of software components
denominated \emph{tasks}, which have dependencies between them, and
impose a partial execution order of tasks. Scientific workflows are
useful for modeling a large number of engineering problems in areas
such as Astronomy~\citet{Ramakrishnan2008}, and Environmental Engineering~\citet{Maechling2007}.
The conjunction of such structure of dependencies and the heterogeneity
of tasks determine that, during the execution of the workflow, workload
can notably vary, which is a scenario suitable for resource elasticity
capabilities of cloud infrastructures~\citet{Monge2017}.

Autoscalers exploit such elasticity capabilities by dynamically adapting
the cloud infrastructure according to the computational workflow requirements.
At the same time, autoscaling must be performed having in mind some
optimization criteria such as execution time, monetary cost or other
aspects of the workflow execution. To such end, autoscalers acquire
or terminate virtual machines~(VMs) and schedule workflow tasks onto
such VMs.

To avoid ambiguity, through the remainder of this article we will
adopt the Amazon Elastic Compute Cloud~(EC2) nomenclature and we
will refer to VMs as \emph{instances. }Instances can be classified
according to: (a)~their \emph{type}, which determines a specific
preset of hardware and software configurations\footnote{EC2 currently offers nearly 60 different instance types.},
and (b)~the \emph{pricing model} adopted, i.e., the on-demand and
the spots models. Instance \emph{types} determine different computing
capabilities and configurations such as the number of CPUs, CPU speed,
memory size, operating system, etc. These characteristics determine
the processing time of executed tasks, in other words, they rule tasks'
performance.

On the other hand, the \emph{pricing} \emph{model} determines the
cost of instances and their behavior. Instances acquired under the
spots model\footnote{Amazon EC2 spot instances: \href{http://aws.amazon.com/ec2/spot/}{http://aws.amazon.com/ec2/spot/},
{[}Online; accessed September-2018{]}}, are usually cheaper than instances acquired under the alternative
on-demand pricing model. However, instances under the spots model
are unreliable as they can fail, thus abruptly terminating the execution
of any running task assigned to the instance. These failures can potentially
impact on workflow execution time (i.e.~makespan) and monetary cost. 

Workflow autoscaling consider such prone-to-failure instances are
a very challenging problem that implies a compromise between the objectives
of makespan, monetary cost and reliability. In other words, this is
a multi-objective minimization problem. In a previous work of our
own~\citet{Monge2017} we tackled scientific workflow autoscaling
using spot instances by solving a makespan minimization problem subject
to budget constraints (i.e. limiting the maximum cost). To cope with
spot instances unreliability, tasks were heuristically scheduled with
the aim of mitigating the negative effects of out-of-bid~(OOB) errors
on the workflow makespan. Such autoscaler is called Spot Instances
Aware Autoscaling~(SIAA).

In other of our works~\citet{Monge2017b} we addressed the multi-objective
minimization problem aiming at reducing makespan, monetary cost and
the potential impact of OOB errors as stated in the previous paragraphs.
However, in such article we focused on a different type of scientific
application called parameter sweep experiments~(PSEs). Just like
workflows, PSEs are applications that comprise a set of tasks to realize
a given experiment, but in PSEs' tasks are \emph{independent} (i.e.
there are no inter-task dependencies) and mostly homogeneous~\citet{Pacini2015},
which determines a completely different and simpler autoscaling scenario.
Such autoscaler is called Multi-objective Evolutionary Autoscaler~(MOEA)
to solve the mentioned problem.

To the best of our knowledge, most cloud autoscaling strategies have
been proposed for the efficient management of workflow applications
mainly subject to deadline constraints~\citet{Cai2017,DeConinck2016,Li2015}.
Moreover, from the revised literature, we have not found works where
the authors minimize the potential impact of OOB errors along with
minimizing both makespan and monetary cost of scientific workflows
in an online scenario as we propose in this work.

Concretely, the key idea of this work is to propose a new autoscaler
to solve the autoscaling of scientific workflows for online scenarios,
i.e., scenarios where the availability of the virtual infrastructure
varies constantly throughout the applications execution. Motivated
by this idea, this article brings in the following contributions:
\begin{itemize}
\item A novel online multi-objective autoscaler and a major extension to
the mathematical formulation of the autoscaler proposed in~\citet{Monge2017b}
for executing scientific workflows. We call this autoscaler Cloud
Multi-objective Intelligence~(CMI). CMI is an autoscaler for scientific
workflows that considers spot instances and minimizes makespan, cost
and the potential impact of OOB errors. The mathematical formulation
of the problem and a description of the CMI autoscaler are given in
Section~\ref{sec:Tri-objective-Autoscaling}.
\item The proper parameterization values for the optimization algorithm
used by CMI considering four real large-scale scientific workflows
derived from the Astronomy and Environmental Engineering~\citet{Juve2013,Ramakrishnan2008}
domains. Tasks execution data extracted from such workflows, actual
characteristics of the Amazon EC2 instances used, on-demand prices,
and data of spot prices are discussed in Section~\ref{sec:Study-cases}.
Then, Section~\ref{sec:NSGA-II-parametrization} presents a sensitivity
analysis performed to obtain the best hyperparameter sets for the
genetic part of CMI.
\item An experimental validation demonstrating that CMI outperforms SIAA~\citet{Monge2017},
a recently published state-of-the-art autoscaler for workflow applications
w.r.t. relevant metrics, particularly makespan, economic cost, and
$L_{2}$-norm of such metrics. As CMI, SIAA also considers spot instances,
but in contra-position, SIAA is a heuristic-based method rather than
a meta-heuristic one. In our experiments, SIAA was parameterized to
operate under different configurations to settle the basis for a rigorous
comparison. Section~\ref{sec:Autoscalers-Comparison} discusses the
mentioned experiments and presents the results of statistical significance
tests that back up the strength of our claims.
\end{itemize}
Finally, Section~\ref{sec:Related-work} surveys relevant related
work and Section~\ref{sec:Conclusions} concludes this work and discusses
future prospective extensions.

\section{\label{sec:Tri-objective-Autoscaling}Online Multi\textendash objective
Workflow Autoscaling}

This work deals with the problem of autoscaling \emph{scientific and
engineering workflows }in public clouds like the Amazon EC2. Autoscaling
workflows requires the acquisition of instances that might belong
to different types and they can be purchased under different pricing
models, in order to optimize a given criteria. In this work, we propose
to minimize workflow makespan, monetary cost and probability of out-of-bid
(OOB) errors. These characteristics make the problem very complex
since it has many possible solutions.

In practice, when a scientific workflow is executed, tasks' \emph{performance}
vary across different instance \emph{types}\footnote{A complete list of the different instance types offered by Amazon
EC2 is available at \url{https://aws.amazon.com/ec2/instance-types/}.}, where each type defines specific characteristics of the instances
to acquire such as processing power, number of processors, memory
size, memory speed, disk size, operating system, etc.

In this work, we assume that instances are charged by hour under two
possible pricing models namely the on-demand model and the spots model.
The \emph{on-demand} pricing model offers the access to reliable instances
at a fixed price charged by hour of computation. Instances acquired
under this pricing model are usually more expensive than instances
acquired using the spots model. For simplicity, we will refer to instances
acquired under the on-demand pricing model as \emph{on-demand instances}.

Alternatively, under the \emph{spots pricing} model, instance's prices
fluctuate over time and are usually much lower than the prices of
the on-demand model. To request an instance under the spots model,
the user must bid the maximum price he/she is willing to pay, and
while the actual price is below the user's bid, the instance will
remain operative. However, if the actual spot price overcomes the
user's bid, an OOB error occurs forcing the termination of the affected
instance and thus the execution of all of the tasks that might be
running on it. Thus, we say that these kind of instances are \emph{unreliable}.
We will refer such instances as \emph{spot instances}.

It is worth to remark that spot-instance prices vary in an unpredictable
way, therefore the occurrence of OOB errors is bound to the bid price
selected. Moreover, considering the large scale of scientific workflows,
the extent of the potential impact of OOB errors depends also on the
amount of spot instances running. For such reason, during the execution
of a scientific workflow, decisions involving spot instances are very
important as they compromise execution makespan, monetary cost and
reliability. For example, lower bids allow to reduce execution costs
but at the same time they increase the probability that an OOB error
occurs, which negatively affects makespan. Conversely, higher bids
reduce the probability of OOB errors, which helps in preserving makespan
but they increase the overall cost of executing a workflow. Therefore,
it is for these reasons that we are dealing with an online multi\textendash objective
problem where it is important to minimize the occurrence of OOB errors,
makespan, and cost~\citet{Monge2017b}. This is a complex problem
and it has not been previously addressed with the same approach adopted
in this work.

Concretely, our autoscaling strategy, called Cloud Multi-objective
Intelligence~(CMI), aims at determining the proper amount of instances
for each combination of instance type and pricing model, and the adequate
bid for spot instances needed for the workflow at hand. The objective
of CMI is to obtain the proper virtual infrastructure configuration
for minimizing the makespan, cost and the potential impact of OOB
errors on the execution of workflow tasks. Moreover, it is important
to mention that autoscaling accounts two interrelated problems:
\begin{enumerate}
\item determining a scaling plan $X$ that describes the adequate virtual
infrastructure setting to request to the cloud provider, i.e. the
amount of instances for each combination of instance type and pricing
model, and the bid price for the accounted spot instances; and
\item scheduling the workflow's tasks onto the instances acquired according
to the scaling plan. 
\end{enumerate}
Both of these problems are NP-hard, and therefore many researchers
have proposed heuristic and meta-heuristic solutions~\citet{Mao2013,Monge2017,Monge2017b}.
Remarkably, our proposed CMI autoscaler searches for solutions which
include the bid price for the spot instances chosen instead of requiring
a particular spot prices prediction method. This is a fundamental
difference with other state-of-the-art autoscalers that use bid prediction
methods~\citet{Monge2017,Turchenko2013}. 

On the other hand, it is also important to remark that devising a
full-ahead scaling plan for the entire workflow execution is impracticable
since it might lead to serious losses on makespan and/or cost. The
reasons are, first, because cloud's performance is variable and therefore
the exact duration of tasks is unpredictable, and second, because
the occurrence of OOB errors is also unpredictable: given bid the
occurrence of OOB errors depends on the future \textendash and unknown\textendash{}
progression of spot prices. 

For the above-mentioned reasons, CMI operates in an online way by
alternating the search of scaling plans on one hand and the scheduling
of tasks on the other hand. In this manner, decisions taken are suited
to the actual progression of the workflow execution and the infrastructure,
which leads to better makespan and cost savings.

CMI, and most autoscalers, operate during the entire execution of
the workflow by periodically solving various autoscaling subproblems
along the time in order to obtain the adequate scaling plan and task
scheduling according to workflow execution progress and cloud infrastructure
load state. Then, in subsections~\ref{subsec:Mathematical-formulation}
and~\ref{subsec:CMI-Autoscaler} we present the mathematical formulation
that defines each one of the autoscaling subproblems that need to
be solved during the workflow's execution, and the CMI autoscaler,
respectively.

\subsection{\label{subsec:Mathematical-formulation}Mathematical Formulation
of Subproblems}

As we discussed before, the autoscaling process adapts the infrastructure
periodically by solving a series of autoscaling subproblems subject
to the current infrastructure state and the execution progress of
the workflow. Each subproblem consist on determining a scaling plan
that specifies the amount of on-demand and spot instances of each
type, as well as the bids for such spot instances. Given $\vmtypes$,
the set of instance types considered for autoscaling, and $n=\left|\vmtypes\right|$
the amount of instance types, we define a scaling plan $\xsca$ is
formally defined as a 3-tuple $(\Xond,\Xspt,\Xbid)$, where:
\begin{itemize}
\item $\Xond=(\xond_{1},\xond_{2},...,\xond_{n})\in\mathbb{N}_{0}^{n}$
is a vector that describes the amount of on-demand instances for each
of the $n$~instance types, each component ranges between 0 and the
maximum number of instances that establishes the cloud provider, 
\item $\Xspt=(\xspt_{1},\xspt_{2},...,\xspt_{n})\in\mathbb{N}_{0}^{n}$
is a vector that describes the amount of spot instances for each of
the $n$~instance types, each component ranges between 0 and the
maximum number of instances that establishes the cloud provider, and
\item $\Xbid=(\xbid_{1},\xbid_{2},...,\xbid_{n})\in\mathbb{R}_{\geq0}^{n}$
is a vector of real values greater or equal to zero that represents
the bid price for the the spot instances described on $\Xspt$, each
value $\xbid_{i}$ is the bid for the $\xspt_{i}$ instances of type
$i$, each component ranges between the actual spot price and the
on-demand price for such instance type.
\end{itemize}
Moreover, given the set of workflow tasks considered in the current
autoscaling step, $\periodtasks$, and the set of available instance
types, $\vmtypes$, a multi-objective autoscaling subproblem is defined
as the minimization of the following three functions showed in Eq.~(\ref{eq:objectiveFuncions}):{\small{}
\begin{equation}
\min\,(\objA(X),\objB(X),\objC(X)),\label{eq:objectiveFuncions}
\end{equation}
}{\small\par}

subject to a set of constraints that restrict the feasible solution
space and define the current status of the execution at the time that
the the subproblem is being solved.

\paragraph{1) Makespan}

The $\objA(\cdot)$ function from Eq.~(\ref{eq:makespan}) represents
the expected execution time of the period tasks $\periodtasks$. To
compute a solution's makespan, the model simulates the behavior of
the scheduling algorithm considering the tasks in $\periodtasks$
and the instances described in $\Xond$ and $\Xspt$. Tasks are scheduled
starting by those with lower margin for delays without violating the
precedence order dictated by the workflow dependencies. Then, for
each task the earliest completion time~(ECT) criterion is used to
select the most appropriate instance. Note that this criterion permits
the assignment of tasks to instances that are not the ones that offer
the shortest execution times but those which promise the earliest
completion times.

Formally, a solution's makespan is computed as:

{\small{}
\begin{equation}
\objA(\xsca)=\underset{t\in\periodtasks}{\max}\{\mathrm{EST}(t)+d_{t}\}-\underset{t\in\periodtasks}{\min}\{\mathrm{EST}(t)\}\label{eq:makespan}
\end{equation}
}{\small\par}

\noindent where~$d_{t}$ is the duration of a task~$t$ which in
practice is mostly unknown and hence should be estimated by using
a performance prediction mechanism~\citet{Monge2015}. Here, durations
are estimated considering the ECT processor from the instances described
in $\Xond$ and $\Xspt$. EST is the earliest start time and accounts
for the minimum time at which a task can start its execution taking
into account its parent tasks, i.e. the preceding tasks according
to workflow dependencies. Moreover, the EST of a waiting task~$t$
is computed as $\mathrm{EST}(t)=\underset{1\leq k\leq p}{\max}\{\mathrm{EST}(t_{k})+d_{k}\},$
where $t$ is a waiting task, $t_{k}$ is one of the $p$ parent tasks
of~$t$ and $d_{k}$ is the estimated duration of~$t_{k}$. Then,
for tasks that are \emph{ready} to execute, the EST is set to the
current time, i.e. when the current autoscaling subproblem is being
solved.

Please note that the makespan function represents a very rugged optimization
hyper-surface because it heavily depends on (a) the heterogeneity
of tasks as the different possible assignments of tasks\textendash instances
provide a wide spectrum of durations, and (b) the workflow dependencies
as they determine different patterns of tasks sequence/parallelism
and propagate delays from parent to children tasks in cascade. Both
of these characteristics, tasks heterogeneity and dependencies between
them, make workflow autoscaling a particularly challenging problem.
This is an important difference with respect to our previous work
in~\citet{Monge2017b} where we had focused on applications that
comprise a set of homogeneous and independent tasks (without dependences
among them), which determined a simpler autoscaling scenario. 

\paragraph*{2) Cost}

The $\objB(\cdot)$ function from Eq.~(\ref{eq:cost}) represents
the cost of running all the instances for one hour of computation. 

{\small{}
\begin{equation}
\objB(\xsca)=\stackrel[i=1]{n}{\sum}\xond_{i}\mathrm{\cdot}\mathrm{price}_{i}+\xspt_{i}\mathrm{\cdot}\xbid_{i},\label{eq:cost}
\end{equation}
}{\small\par}

In the case of on-demand instances, the model computes the total cost
of executing all the instances of type $i$ ($\xond_{i}$) multiplied
by their corresponding on-demand price ($\mathrm{price}_{i}$). In
the case of spot instances, the model estimates the cost in a different
way. As the price of spot instances is determined by the cloud provider
in a unpredictable way, the model estimates the cost by multiplying
the amount of instances of type $i$ ($\xspt_{i}$) by their corresponding
bid ($\xbid_{i}$). In this estimation we also assume that OOB errors
do not occur. 

Note these two assumptions regarding the cost of spot instances lead
in our approach to a pessimistic estimation of the overall cost because
of the following two reasons. First, as OOB errors are not considered
(although they could indeed occur during the actual execution) the
cost of the hypothetically failing instances will be added up to the
overall cost. Second, for each instance we used the bid as the spot
price which is an upper bound of the actual spot prices.

\paragraph*{3) Potential impact of OOB errors}

The $\objC(\cdot)$ function from Eq.~(\ref{eq:error-probability})
estimates the potential impact on the workflow considering the amount
of spot instances and their associated bids. As the occurrence of
OOB errors depends on the chosen bid and the future evolution of spot
prices, it cannot be exactly determined when and for which bids, OOB
errors will occur. For such reason, and to cope with this source of
unavoidable uncertainty, our formulation uses a probabilistic function
of OOB-error occurrences. Formally, the OOB errors are computed as:

{\small{}
\begin{equation}
\objC(\xsca)=\stackrel[i=1]{n}{\sum}\xspt_{i}\cdot\mathrm{vCPU_{\mathit{i}}}\cdot P_{i}(\xbid_{i}),\label{eq:error-probability}
\end{equation}
}{\small\par}

To estimate the potential impact of OOB errors derived from choosing
a given bid ($\xbid_{i}$) for the amount of instances of type $i$
($\xspt_{i}$), the $\objC(\cdot)$ function computes a weighted sum
of the OOB error probabilities $P_{i}(\xbid_{i})$. The probability
of OOB errors for each instance type $i$ given the selected bid,
$P_{i}(\xbid_{i})$, is multiplied by the maximum number of tasks
that could be executing on the instances of such type. As a processor
of any instance can not execute more than one task at a time, the
maximum number of tasks that might be executed on the instances of
the $i$\textsuperscript{th} type is simply the total amount of processors
of the spot instances such type, formally, $\xspt_{i}\cdot\mathrm{vCPU_{\mathit{i}}}$.
In this way the function estimates the potential impact related to
the number of tasks which can fail due to an OOB error. The objective
function computes the overall impact of OOB errors considering all
the instance types.

\paragraph*{Constraints}

The mathematical formulation for each subproblem also includes the
following constraints: 

{\small{}
\begin{equation}
\objB(X)\leq\mathrm{B},\label{eq:budget-constraint}
\end{equation}
}{\small\par}

{\small{}
\begin{equation}
\mathrm{X}_{i}^{\emph{{min}}}\leq x_{i}^{\mathrm{\emph{{od}}}}+x_{i}^{\mathrm{s}}\leq\mathrm{X}_{i}^{\mathrm{\emph{{max}}}},\label{eq:instance-constraints}
\end{equation}
}{\small\par}

{\small{}
\begin{equation}
\stackrel[i=1]{n}{\sum}\xond_{i}+\xspt_{i}>0,\label{eq:minimum-instances-constraint}
\end{equation}
}{\small\par}

{\small{}
\begin{equation}
\mathrm{S_{i}^{\emph{{currentPrice}}}\leq\xbid_{i}\leq\mathrm{price}_{i},}\label{eq:bid-constraints}
\end{equation}
}{\small\par}

\noindent where $\mathrm{X}_{i}^{\mathrm{min}}$ and $\mathrm{X}_{i}^{\mathrm{max}}$
are, respectively, the minimum and maximum number of allowed instances
for the $i$\textsuperscript{th} instance type, $\mathrm{S}_{i}^{\mathrm{currentPrice}}$
is the current price for the spot instances of type $i$, $\mathrm{price}_{i}$
is the on-demand price for the $i$\textsuperscript{th} instance
type. The minimum amount of instances~($\mathrm{X}_{i}^{\mathrm{min}}$)
is the amount of running instances that are processing at least one
task for each instance type as we can not terminate instances that
are executing tasks.

The budget constraint in Eq.~(\ref{eq:budget-constraint}) limits
the cost of the current autoscaling solution to be below the maximum
monetary budget $B$. The $n$ constraints represented by Eq.~(\ref{eq:instance-constraints})
determine the minimum and {\small{}maximum} number of instances for
each type. The constraint in Eq.~(\ref{eq:minimum-instances-constraint})
forces that autoscaling solutions must consider at least one instance.
Finally, the $n$ constraints in Eq.~(\ref{eq:bid-constraints})
circumscribe bids to be between the current spot price, $\mathrm{S}_{i}^{currentPrice}$,
and a maximum spot price, $\mathrm{S}_{i}^{max}$, for each instance
type.

\subsection{\label{subsec:CMI-Autoscaler}CMI Autoscaler}

\begin{figure}[h]
\begin{centering}
\includegraphics[width=1\textwidth]{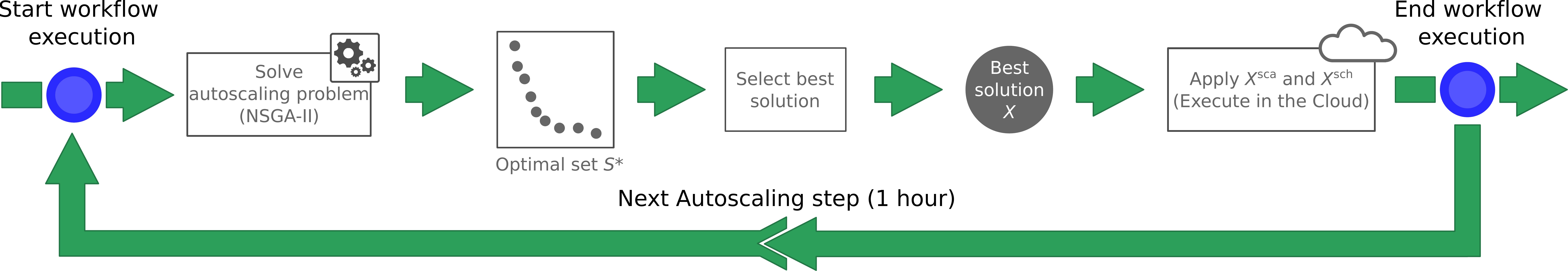}
\par\end{centering}
\caption{\label{fig:autoscaling-process}Illustration of the complete autoscaling
process of CMI.}
\end{figure}

During the execution of a given workflow, the CMI autoscaler iteratively
solves different cases of the multi\textendash objective subproblems,
which are defined by the current state of the workflow execution (i.e.
the instances running, the degree of advance of tasks, and the current
bid prices). In order to solve each autoscaling subproblem, CMI exploits
the well\textendash known multi-objective evolutionary algorithm called
Non-dominated Sorting Genetic Algorithm~II~(NSGA-II)~\citet{Deb2002}.
CMI finds a set of Pareto-optimal solutions and then chooses the one
that best fulfills a pre-established selection criterion. Then, CMI
applies the scaling plan corresponding to such solution and schedules
the tasks accordingly. Figure~\ref{fig:autoscaling-process} illustrates
the complete autoscaling process carried out by CMI.

From the figure it can be seen that at each autoscaling step, CMI
goes through three phases:
\begin{enumerate}
\item solving the multi-objective autoscaling subproblem at hand using the
NSGA-II algorithm to obtain a set of Pareto-optimal solutions~$\mathcal{S}^{*}$, 
\item selecting a solution $\vect x$ according to a user\textendash defined
criterion, and
\item constructing and applying the resulting scaling plan $\xsca$ and
schedules tasks.
\end{enumerate}

\subsubsection{Solving the Multi-objective Subproblem}

NSGA-II~\citet{Deb2002} is a popular multi\textendash objective
evolutionary algorithm that has been applied to many real world problems~\citet{Montoya2014,Rekik2017,Yu2015}.
NSGA-II initializes the population randomly according to the variable
ranges for the given autoscaling subproblem. Then, the population
is sorted based on non-domination in a series of fronts. The first
front is a completely non-dominant set for the current population
and the second front is dominated by the individuals only in the first
front, and so on. The fitness of each individual corresponds to number
of front it belongs to. Crowding distance is also calculated for each
individual. Such distance measures the distance between an individual
and its neighbors. Parents are selected from the population by using
binary tournament based on both, fitness and crowding distance, which
promotes the selection of non-dominated diverse solutions. The selected
parents produce an offspring by applying the Simulated Binary Crossover~(SBX)
operator, which attempts to simulate the offspring distribution of
binary-encoded single-point crossover on real-valued variables. Resulting
individuals are also mutated by applying the Polynomial Mutation~(PM)
operator, which attempts to simulate the offspring distribution of
binary-encoded bit-flip mutation on real-valued variables. These two
operators favor offspring nearer to the parent.

Individuals in the current population and current offsprings are sorted
based on non-domination and only the best $k$ individuals are selected
based on fitness and crowding distance as indicated earlier. In this
context, $k$ is the population size. Finally, the algorithm terminates
its execution once the maximum number of evaluations for the generated
individuals is met, returning the non-dominated solutions obtained
so far.

\subsubsection{\label{subsec:Solution-selection}Selection of the Best Solution}

From the solutions in the optimal set $\mathcal{S}^{*}$ obtained
by NSGA-II, CMI applies a global criterion method to select one of
the candidate solutions. This strategy permits obtaining a solution
in an autonomous way without relying on an human decision maker. The
adopted selection criterion is the following: $\mathrm{arg}\min_{\vect x}\left\Vert f(\vect x)-\zideal\right\Vert $,
where $\vect x\in\mathcal{S}^{*}$. This method selects the solution
$\vect x$ that minimizes the distance to an ideal solution $\zideal$,
where $\left\Vert \cdot\right\Vert $ is the $L_{2}$ norm. We choose
$\zideal=(0,0,0)$ as the ideal solution, which corresponds to the
solution with failure probability, makespan and cost equal to 0. This
policy tends to select the most balanced solution considering the
values for each one of the objectives.

\subsubsection{Scaling Plan and Task Scheduling}

Based on the selected solution $\vect x$, the autoscaler reconstructs
the scaling plan. The autoscaler acquires the amount of on-demand
($\Xond$) and spot instances ($\Xspt$) with the corresponding bid
prices ($\Xbid$) indicated in the selected solution $\vect x$. Then, tasks
in $\periodtasks$ are scheduled using the ECT policy on the requested
instances acquired according to the scaling plan.

For validating the described autoscaler we performed a comparative
study against a state-of-the-art autoscaler called SIAA~\citet{Monge2017}
over four real world workflow applications from the Astronomy and
Environmental Engineering domains. Next section describes the main
characteristics of these applications.

\section{\label{sec:Study-cases}Study Cases}

This Section describes four scientific workflows that serve as different
study cases due to their own particular characteristics, mainly in
terms of: (\emph{i}) the ranges of task durations \textendash from
a few minutes of computation in some workflows to days in others\textendash ,
and (\emph{ii}) the structure of workflow dependencies, which determine
very different patterns and possibilities of task execution parallelism.

These workflows come from the areas of Astronomy and Environmental
Engineering~\citet{Juve2013,Ramakrishnan2008} and were used to validate
the performance of the autoscaling strategies under realistic workload
patterns. The scientific workflows are described below:
\begin{description}
\item [{CyberShake}] CyberShake is a workflow used in the Southern California
Earthquake Center (SCEC)~\footnote{Southern California Earthquake Center: \url{http://www.scec.org}}
for characterizing hazards generated by earthquakes. CyberShake performs
a Probabilistic Seismic Hazard Analysis~(PSHA) over a geographic
region. Message Passing Interface~(MPI) tasks performing finite difference
simulations are executed to generate Strain Green Tensors (SGTs).
The generated SGT data is used to compute synthetic seismograms for
each of the predicted ruptures. Finally, spectral acceleration and
probabilistic hazard curves are generated. 
\item [{LIGO's~Inspiral}] Laser Interferometer Gravitational Wave Observatory~(LIGO)~\footnote{LIGO's Inspiral: \url{http://www.ligo.caltech.edu/}}
is in the crusade for the detection of gravitational waves resulting
from several events in the universe according to Einstein's theory
of general relativity. The LIGO Inspiral Analysis Workflow (LIGO's
Inspiral) analyzes data from the coalescing of binary neutron stars
and black holes. Time-frequency data from events captured by each
of the three LIGO detectors is split into smaller chunks for analysis.
Each chunk is used to generate a subset of waveforms within a given
parameter space later used for computing a matched filter output.
When a true inspiral is detected a trigger is generated and it is
checked against triggers from the other detectors. From now on we
will refer to this workflow simply as \Lig{}.
\item [{Montage}] Montage is a toolkit created by the Infrared Processing
and Analysis Center~(IPAC)\footnote{IPAC: \url{http://www.ipac.caltech.edu/}}
of the NASA~\footnote{NASA: \url{http://www.nasa.gov/}} aimed to
generate custom mosaics of the sky from a set of input images. The
geometry of such input images is used to compute the geometry of the
final mosaic. The geometry is used to \emph{re-project} the images
into the same scale and rotation. Then, the images are corrected for
standardizing the different background emissions to the same level.
Re-projected and corrected images are added together into the final
mosaic. 
\item [{Pan-STARRS}] Panoramic Survey Telescope And Rapid Response System
(Pan-STARRS's) project~\footnote{Pan-STARRS: \url{http://www.ifa.hawaii.edu/}}
is a continuous survey of the entire sky. The data are collected by
a telescope and used to detect hazardous objects in the Solar System,
and other astronomical studies including cosmology and Solar System
astronomy. The astronomy data from Pan-STARRS is managed through PSLoad
and PSMerge workflows. The PSLoad stages incoming data files from
the telescope pipeline and loads them into individual relational databases
each night. Periodically the online production databases that can
be queried by practitioners, are updated with the databases collected
over the week by a second workflow called PSMerge. Both PSLoad and
PSMerge workflows are data intensive but require coordination and
orchestration of resources to ensure reliability and integrity of
the data products. 
\end{description}
These workflows present differences on their structure and alternate
phases of parallel and sequential tasks, which in principle require
a timely and precise scaling up or down of the cloud infrastructure
to properly reduce makespan and cost. An additional aspect to take
into consideration in this analysis is the duration of tasks that,
in conjunction with the mentioned structural characteristics of the
workflow, will determine the preference over parallel/sequential execution
of tasks, predominance of on-demand versus spot instances, and selection
of higher or lower bid prices to request spot instances. All these
factors shape a complex spectrum of execution conditions that an autoscaler
must consider to properly conduct the execution.

Note that these different workload and parallelism patterns condition
the possibilities of instances acquisition and tasks scheduling while
trying to minimize makespan, cost and number of OOB errors. For example
multiple parallel short-duration tasks are best suitable for spot
instances at low bids as the potential of cost and makespan reduction
is large and subject to a low instance failure probability. Conversely,
long\textendash duration critical tasks are best suitable for on-\-demand
instances, which are more reliable (as they are not subject to OOB
errors) but are more expensive.

\section{\label{sec:NSGA-II-parametrization}NSGA-II Parameterization}

\begin{figure}[h]
\begin{centering}
\includegraphics[width=1\textwidth]{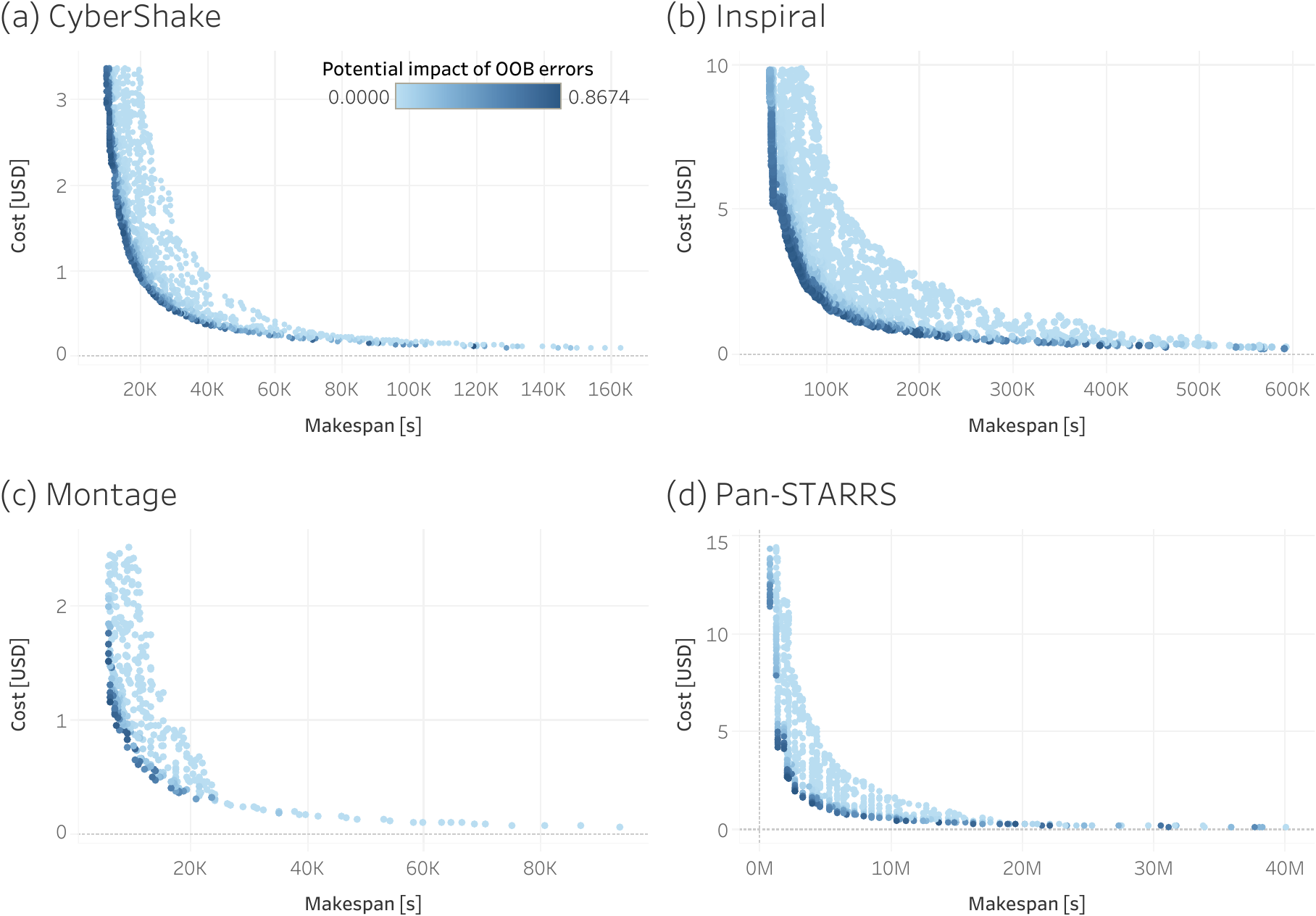}
\par\end{centering}
\caption{\label{fig:pareto-fronts}Combined 3D Pareto fronts for each workflow
obtained by NSGA-II. Fronts were obtained by the combination of solutions
obtained for all the parameter samples (140) and seeds (30). Potential
impact of OOB errors are presented by the intensity of color.}
\end{figure}

In order to obtain the best NSGA-II parameterization for each studied
workflow we performed a sensitivity analysis by sampling~140 parameter
configurations using the Saltelli sampling method~\citet{Saltelli2008}
and performing~30 repetitions for each configuration using different
random number generator seeds. The following list presents the NSGA-II
parameter exploration ranges:
\begin{itemize}
\item \emph{maxEvaluations} (max eval.): the number of solutions to be evaluated
before terminating the algorithm. Sampling interval: $[500,10000]$,
step 1.
\item \emph{populationSize} (pop. size): the number of solutions to keep
across generations. Sampling interval: $[100,400]$, step 1.
\item \emph{sbx.rate} (SBX rate): the probability of applying the Simulated
Binary Crossover~(SBX) on a pair of individuals. Sampling interval:
$[0.8,0.9]$, step 0.01.
\item \emph{sbx.distributionIndex} (SBX dist.): The distribution index controls
the shape of the offspring distribution. Larger values for the distribution
index generates offspring closer to the parents. Sampling interval:
$[5.0,20.0]$, step 0.01.
\item \emph{pm.rate} (PM rate): the probability of applying the Polynomial
Mutation (PM) on each individual. Sampling interval: $[0.5,0.6]$,
step 0.01.
\item \emph{pm.distributionIndex} (PM dist.): As in the case of SBX, the
distribution index controls the shape of the offspring distribution.
Larger values for the distribution index generates offspring closer
to the parents. Sampling interval: $[10.0,15.0]$, step 0.01.
\end{itemize}
Figure~\ref{fig:pareto-fronts} presents the combined Pareto fronts
resulting from the samplings and seeds used for each workflow. For
the sake of clarity, scatter pots exclude extreme solutions, i.e.
those solutions with extreme high cost or high makespan. Note that
we can allow us this simplification as the aim of our autoscaler is
to produce balanced solutions regarding the modeled problem objectives.
As the solution method selector described in Section~\ref{subsec:Solution-selection}
chooses solutions closer to (0, 0, 0), we can focus on the portion
of the Pareto-fronts closer to the origin.

From all the evaluated parameter samples and repetitions we applied
the Sobol method for variance-based sensitivity analysis~\citet{Tang2006}.
In general terms, this method decomposes the variance of a model's
output into portions associated to the sets of inputs for such model.
In the context of this study, we are interested in analyzing the effect
of the NSGA-II parameters on the variance of the NSGA-II performance.
Therefore we choose the hypervolume~(HV) metric of the obtained Pareto\textendash fronts
as the performance indicator, as it captures the degree to which obtained
solutions are close to a reference set and how they are distributed.

The Sobol method is useful because it permits measuring the sensitivity
of NSGA-II across the space of parameters discriminating the effects
of different subsets of parameters. In this study we focus on the
analysis of total\textendash order effects, which measure the contribution
of each parameter to the output variance, which includes the variance
caused by any of its interactions of any order with any other parameter.
In this context, order indicates the number of variables considered
in the interactions, first\textendash order refers to the effects
of a single variable, second order refers to the effects of pairs
of variables and so on. Table~\ref{tab:Sobol-sensitivity} presents
the results of the sensitivity analysis for each workflow.
\begin{table}[h]
\caption{\label{tab:Sobol-sensitivity}Sensitivity analysis results for NSGA-II
discriminated by workflow. Values represent the total\textendash order
effects of each of the parameters NSGA-II parameters: maxEvaluations
(max eval.), populationSize (pop. size), sbx.rate (SBX rate), sbx.distributionIndex
(SBX dist.), pm.distributionIndex (PM dist.), pm.rate (PM rate) and
pm.distributionIndex (PM dist.). Parameters max eval. and pop. size
are the most sensitive ones for every studied workflow.}

\begin{centering}
{\small{}}{\small\par}
\par\end{centering}
\centering{}%
\begin{tabular*}{1\textwidth}{@{\extracolsep{\fill}}lcccccccc}
\toprule 
\multirow{2}{*}{Workflow} &  & \multirow{2}{*}{max eval.} & \multirow{2}{*}{pop. size} & \multicolumn{2}{c}{SBX} &  & \multicolumn{2}{c}{PM}\tabularnewline
\cmidrule{5-6} \cmidrule{8-9} 
 &  &  &  & rate & dist. &  & rate & dist.\tabularnewline
\midrule
\midrule 
\Cyb &  & \textbf{1.141} & \textbf{0.123} & -0.025 & -0.008 &  & 0.006 & -0.044\tabularnewline
\Lig &  & \textbf{1.104} & \textbf{0.099} & -0.033 & 0.020 &  & 0.033 & 0.091\tabularnewline
\Mon &  & \textbf{1.187} & \textbf{0.151} & 0.019 & 0.064 &  & 0.099 & 0.124\tabularnewline
\Pan &  & \textbf{1.139} & \textbf{0.129} & -0.027 & -0.042 &  & -0.050 & -0.123\tabularnewline
\bottomrule
\end{tabular*}
\end{table}

From the table it can be seen that max eval. and pop. size are the
most sensitive parameters (indicated by the large values). These results
indicate that the different values that such parameters might take
have strong effect on the resulting HV values, i.e. on the performance
of NSGA-II. In other words, selecting inappropriate values for such
parameters will inhibit NSGA-II from obtaining good quality solution
sets.

The lower sensitivity values taken by the remaining parameters indicate
that in terms of relevance their effects are smaller than the effects
of the first cited parameters. Nonetheless, for the proper operation
of NSGA-II, it is also important to determine the right values of
such parameters.

With the aim of obtaining the best parameterization of NSGA-II for
the autoscaler we choose the one that maximizes the HV metric. By
maximizing the HV we ensure that the parameters set selected promise
a good level of convergence and diversity on the solutions with respect
to the reference sets. Table~\ref{tab:best-parameters} reports the
selected NSGA-II parameterization and the corresponding HV value for
each workflow.
\begin{table}[h]
\caption{\label{tab:best-parameters}Selected parameters for each workflow.}

\centering{}%
\begin{tabular*}{1\textwidth}{@{\extracolsep{\fill}}lcccccccccc}
\toprule 
\multirow{2}{*}{Workflow} &  & \multirow{2}{*}{max eval.} & \multirow{2}{*}{pop. size} & \multicolumn{2}{c}{SBX} &  & \multicolumn{2}{c}{PM} &  & \multirow{2}{*}{HV}\tabularnewline
\cmidrule{5-6} \cmidrule{8-9} 
 &  &  &  & rate & dist. &  & rate & dist. &  & \tabularnewline
\midrule
\midrule 
\Cyb &  & 9712 & 145 & 0.88 & 18.9 &  & 0.55 & 10.8 &  & \textbf{0.994}\tabularnewline
\Lig &  & 9712 & 145 & 0.82 & 17.2 &  & 0.58 & 12.5 &  & \textbf{0.997}\tabularnewline
\Mon &  & 9712 & 145 & 0.88 & 18.9 &  & 0.55 & 10.8 &  & \textbf{0.996}\tabularnewline
\Pan &  & 9712 & 145 & 0.88 & 18.9 &  & 0.55 & 10.8 &  & \textbf{0.990}\tabularnewline
\bottomrule
\end{tabular*}
\end{table}

Note that for the most sensitive parameters, max eval. and pop. size,
the selected configurations include those values closer to the upper
bound of the parameterization ranges presented at the beginning of
this section. These values make the genetic algorithm to perform the
search working on a larger pool of solutions and letting those solutions
evolve through a larger number of generations, which leads to the
obtainment of better sets of candidate solutions for the autoscaling
problems. These parameters are used in the comparative study presented
in the next section.

\section{\label{sec:Autoscalers-Comparison}Comparative study}

In order to evaluate the performance of CMI we compare it with another
state-of-the-art autoscaler named Spot Instances Aware Autoscaling~(SIAA),
which is described in subsection~\ref{subsec:SIAA-autoscaler}. Then,
details on the experimental settings and experimental results are
provided in subsections~\ref{subsec:Experimental-Settings} and~\ref{subsec:Experimental-Results},
respectively. Finally, an analysis of statistical significance of
results is provided in subsection~\ref{subsec:Statistical-significance}.

\subsection{\label{subsec:SIAA-autoscaler}Spot Instances Aware Autoscaling~(SIAA)}

SIAA~\citet{Monge2017} autoscaler starts determining a scaling plan
and then schedules the tasks for running on the available instances.
To this end, SIAA uses, on the one hand, a scaling algorithm which
permits the acquisition of a cloud infrastructure comprising on-demand
and spot instances, and on the other hand, an heuristic scheduling
algorithm to minimize the workflow makespan and reduce the probability
that failing spot instances might interrupt the execution of longer
tasks. In the \emph{scaling phase}, SIAA determines a consumption
vector and then scales it to meet the budget constraint. To such end,
the available budget is split in two parts and assigned to acquire
the maximum amount of on-demand and spot instances according to such
budget distribution. To do this, SIAA uses bidding methods to request
spot instances. 

In this work SIAA uses different configurations of the probabilistic
bidding method discussed on~\citet{Monge2017}, which aim at bidding
prices that promise bid failures below a given probability threshold
that we call \emph{confidence level}. Then, in the \emph{scheduling
phase}, the algorithm tackles the minimization of the workflow makespan
by reducing the execution time of critical tasks. Moreover, the slack
times (the maximum amount of time that a task can be delayed without
delaying any of their child tasks) are used to determines a priority
for the selection of tasks during scheduling. The tasks which are
ready to execute are sorted by their slack times in an increasing
order (first those tasks which have smaller margin for delay). This
criterion executes first those tasks which potentially have more impact
on the workflow makespan. 

The algorithm schedules the tasks \textendash one by one\textendash{}
until there are no more ready tasks to schedule or all the instances
are busy. Tasks are scheduled to instances according to two criteria.
First, tasks are allocated to on-demand instances, but if there are
no available instances, the algorithm checks the availability of spot
instances. This policy favors that failures, if occur, affect non-critical
tasks that could be re-scheduled without a large impact on the overall
makespan as they have a wider margin for delays (larger slack times).
Second, tasks are allocated to the instances that promise the ECT.
This policy is applied while selecting an on-demand or a spot instance
for the task at hand. In this way the algorithm promotes the execution
of tasks on the fastest possible instances.

\subsection{\label{subsec:Experimental-Settings}Experimental Settings}

In order to perform the experiments we instantiated the CloudSim toolkit~\citet{Calheiros2011},
which is heavily used within the community for the design of realistic
cloud experiments. The characteristics and the on-demand price of
the instances considered during the experimentation are presented
in Table~\ref{tab:on-demand-instances} along with their corresponding
on-demand price. Values in the table correspond to real characteristics
of Amazon Elastic Compute Cloud (EC2) instances, where vCPU is the
number of (virtual) CPUs available for each one of the instance types,
ECU\textsuperscript{tot} (acronym for \emph{E}C2 \emph{C}ompute \emph{U}nits)
is the relative computing power of the instances considering all its
virtual CPUs\footnote{One ECU is equivalent to a CPU capacity of a 1.0-1.2~GHz 2007 Opteron},
ECU is the relative performance of each one of the CPUs, and the last
column shows the price in United States dollars~(USD) per hour of
computation. These instance types were selected to provide diverse
spectrum of performance and price configurations.
\begin{table}[h]
\caption{\label{tab:on-demand-instances}Characteristics of the on-demand instances.
The data corresponds to instances belonging to the US-west (Oregon)
region.}

\centering{}%
\begin{tabular*}{1\columnwidth}{@{\extracolsep{\fill}}lcccc}
\toprule 
Instance type & vCPU & ECU\textsuperscript{tot} & ECU & Price (USD)\tabularnewline
\midrule
\midrule 
t2.micro & 1 & 1 & 1 & \$0.013\tabularnewline
m3.medium & 1 & 3 & 2 & \$0.07\tabularnewline
c3.2xlarge & 8 & 28 & 3.5 & \$0.42\tabularnewline
r3.xlarge & 4 & 13 & 3.25 & \$0.35\tabularnewline
m3.2xlarge & 8 & 26 & 3.25 & \$0.56\tabularnewline
\bottomrule
\end{tabular*}
\end{table}

For experimentation we used an actual history of Amazon EC2 spot prices
that corresponds to the period between March 7\textsuperscript{th}
and June 7\textsuperscript{th} of 2016 for the US-west region~(Oregon)~\citet{SpotsData2016}.

The first two months of data were used to compute the OOB errors probabilities
by counting the number of times a sliding time window of 1 hour presented
spot prices over the set of bid values. Note that these bid values
correspond to the bid levels defined in our mathematical formulation
of the scheduling problem. The data corresponding to the last month
of the history was kept for the experiments presented on Section~\ref{sec:Autoscalers-Comparison}
as the spot price variations during the simulation. Using the data
in such way allowed us to perform the evaluation of the bidding methods
and therefore the performance of the autoscaling strategies while
completely ignoring future evolution of spot prices, which is the
real scenario for any EC2 user.

We configured our competitor, SIAA, by varying the proportion of spots
and on-demand instances, i.e. spots ratio~(SR) and the bidding method
confidence~(BMC), which is the bid price that probability of OOB
errors the same bidding method used in~\citet{Monge2017}. The values
that SR can take are $0.1,0.2,...,1.0$ and the values taken by BMC
are $0.05,0.1,...,0.25$, which gives a total of 55~different configurations
of SIAA (11 SR values $\times$ 5 BMC values). To ensure the robustness
of results we repeated (30 times) the simulations of CMI, and SIAA
for each of the 55~configurations, giving a total of 30~executions
of CMI and~1650 of SIAA for each workflow. These configurations were
used for each of the~4 studied workflows giving a total of $30\times4+1650\times4=6720$
simulations.

\subsection{\label{subsec:Experimental-Results}Experimental Results}

Table~\ref{tab:summary} presents the mean values for the number
of on-demand and spot instances, makespan in seconds, cost in USD,
number of task failures, and the $L_{2}$-norm considering the task
failures, makespan and cost resulting from the simulation, which
provides a joint analysis of the metrics of interest for this autoscaling
problem.
\begin{table}[h]
\caption{\label{tab:summary}Summary of results. Each row presents the mean
values per strategy considering all the studied workflows. Except
for the average number of on-demand and spot instances that are merely
descriptive, for the remaining metrics lower values represent better
results. }

\centering{}%
\begin{tabular*}{1\textwidth}{@{\extracolsep{\fill}}lc>{\centering}p{0.11\textwidth}>{\centering}p{0.09\textwidth}cc>{\centering}p{0.09\textwidth}>{\centering}p{0.1\textwidth}}
\toprule 
Workflow & Strategy & No. of on-demand & No. of spot & Makespan & Cost & Task Failures & $L_{2}$ norm\tabularnewline
\midrule
\midrule 
CyberShake & CMI & 6.37 & 85.7 & 18604.02 & \textbf{8.40} & 47.33 & \textbf{0.22}\tabularnewline
 & SIAA & 7.08 & 22.13 & \textbf{18419.21} & 13.62 & 33.13 & 0.71\tabularnewline
\midrule 
Inspiral & CMI & 10.59 & 133.26 & \textbf{108958.56} & \textbf{65.38} & 110.41 & \textbf{0.14}\tabularnewline
 & SIAA & 24.26 & 51.88 & 109494.01 & 155.21 & 135.75 & 0.72\tabularnewline
\midrule 
Montage & CMI & 4.72 & 95.62 & \textbf{29962.68} & \textbf{4.47} & 74.66 & \textbf{0.10}\tabularnewline
 & SIAA & 6.89 & 13.37 & 35806.54 & 8.51 & 30.52 & 0.80\tabularnewline
\midrule 
Pan-STARRS & CMI & 73.52 & 712.12 & 2215656.73 & \textbf{3351.61} & 1579.68 & \textbf{0.39}\tabularnewline
 & SIAA & 49.06 & 105.32 & \textbf{1733431.72} & 6206.71 & 397.23 & 0.63\tabularnewline
\bottomrule
\end{tabular*}
\end{table}

From the table it can be seen that CMI tends to use a larger number
of spot instances than SIAA. Considering makespan, and in the case
of CyberShake and Montage, CMI obtains better average performance
than SIAA. The opposite is true for the remaining workflows. In terms
of cost, CMI outperforms SIAA for all the workflows and with a wide
margin of savings (around 50\% in some cases). Note that the number
of task failures is also much higher for CMI than for SIAA, which
is explained by the larger number of spot instances used and the low
bid prices used. The negative impact of such high number of task failures
is evident for \Lig{} and \Pan{}, which are the workflows with many
very large tasks. However, a very interesting observation for the
remaining workflows is that despite the high number of failures, makespan
is shorter for CMI than for SIAA. The reason is \Cyb{} and \Mon{}
have many short\textendash duration tasks, and if failures occur they
do not heavily affect many tasks, even more, tasks completed in the
last hour of execution of a failing spot instance are executed for
free according to Amazon spot policies. 

Finally the $L_{2}$-norm metric favors CMI in all the cases, which
indicates that such autoscaler outperforms SIAA always in terms of
cost and in two of the studied workflows in terms of makespan. For
the remainder of this section we focus on the\emph{ }$L_{2}$-norm
metric, which provides a joint measure of performance facilitating
the comparison procedure. Analyzing makespan or cost in isolation
does not capture the trade\textendash off that occurs between those
metrics, thus limiting the depth of the derived conclusions.

We can conclude that CMI is not restricted by any proportion of spot
and on-demand instances nor by the bid prices to use, which permits
acquiring different distributions of instances depending on the next
workload pattern. In addition, CMI also determines the bid price to
use for each instance type, which permits adjusting the relation between
the maximum price to pay and reliability according to the upcoming
tasks.

\subsubsection*{Analysis of the Autoscaling Process}

In this section we provide an analysis of the events that occur during
the autoscaling process on a particular simulation scenario (i.e.
same workflow and repetition number). We analyzed four autoscaling
processes for the Montage workflow, one of them corresponding to CMI,
and the others to SIAA with three different configurations. The first
case corresponds to the process carried out by the best-performing
CMI (CMI-b). Then, the remaining cases correspond to three of the
55 configurations of SIAA with: its \emph{best} performing configuration
(SIAA-b), a configuration that produced an \emph{expensive} execution
(SIAA-e), and with a configuration (SIAA-u) for which the execution
is strongly affected by the \emph{unreliability} of spot instances.
Table~\ref{tab:Autoscaling-metrics} summarizes the performance metrics
considering number of instances, virtual CPUs (vCPUs), makespan, cost,
number of task failures and the $L_{2}$-norm metric for the analyzed
cases.
\begin{table}[h]
\caption{\label{tab:Autoscaling-metrics}Metrics for the analyzed cases.}

\centering{}%
\begin{tabular*}{1\textwidth}{@{\extracolsep{\fill}}lcccccc}
\toprule 
Autoscaler & Instances & vCPUs & Makespan & Cost & OOB errors & $L_{2}$-norm\tabularnewline
\midrule
\midrule 
CMI-b & 301 & 450 & \textbf{29268.33} & \textbf{3.21} & \textbf{0} & \textbf{0.045}\tabularnewline
\midrule 
SIAA-b & 26 & 168 & 29991.00 & 4.00 & \textbf{0} & 0.051\tabularnewline
SIAA-e & 26 & 168 & 29991.00 & 22.78 & \textbf{0} & 0.196\tabularnewline
SIAA-u & 27 & 176 & 38648.00 & 4.22 & 1 & 0.060\tabularnewline
\bottomrule
\end{tabular*}
\end{table}

In Figure~\ref{fig:Autoscaling-analysis} we present the autoscaling
processes for the mentioned cases. Curves indicate the number of tasks
and vCPUs through time, and labels indicate the accumulated cost of
execution on each autoscaling step. From the figure it can be seen
that:
\begin{enumerate}
\item \emph{CMI-b} achieves the best performance with respect to makespan
and cost, and therefore also achieves the best performance with respect
to the $L_{2}$-norm~(see Table~\ref{tab:Autoscaling-metrics}).
Moreover, it can also be seen that CMI-b requests a greater number
of VMs than the other three processes. However, this does not impact
the cost because CMI-b chooses mostly spot instances and makes sure
to take bids that do not produce OOB errors.
\item \emph{SIAA-b} (SIAA with the best performing configuration) presents
a good balance between makespan and cost. As can be seen, SIAA-b is
only outperformed by CMI-b. Concretely, the gains of CMI-b with respect
to best-SIAA-b for the makespan and cost are equal to 2.41\% and 19.75\%,
respectively. 
\item \emph{SIAA-e} (SIAA with a configuration that produced an expensive
execution) achieves the same makespan obtained by SIAA-b, however,
it is almost 6 times more expensive than SIAA-b due to the fact that
SIAA-e uses a large number of on-demand instances. Here, the gains
of CMI-b with respect to SIAA-e are equal to 2.41\% and 85.90\% for
the makespan and cost, respectively.
\item \emph{SIAA-u} (SIAA with a configuration for which the execution is
strongly affected by the \emph{unreliability} of spot instances) has
the largest makespan in this analysis. In this case, an OOB error
occurs producing the failure of a critical task, which delays their
termination in detriment of workflow makespan. This event is illustrated
with a vertical red line in the Figure~\ref{fig:Autoscaling-analysis}.
The obtained makespan is considerably greater than CMI-b and the other
SIAA configurations. However, as can be seen in the Figure~\ref{fig:Autoscaling-analysis},
the cost obtained by SIAA-u is quite close to SIAA-b. Finally, the
gains obtained by CMI-b with respect to SIAA-u are equal to 24.27\%
and 23.93\% with respect to the makespan and cost, respectively.
\begin{figure}
\begin{centering}
\includegraphics[width=1\columnwidth]{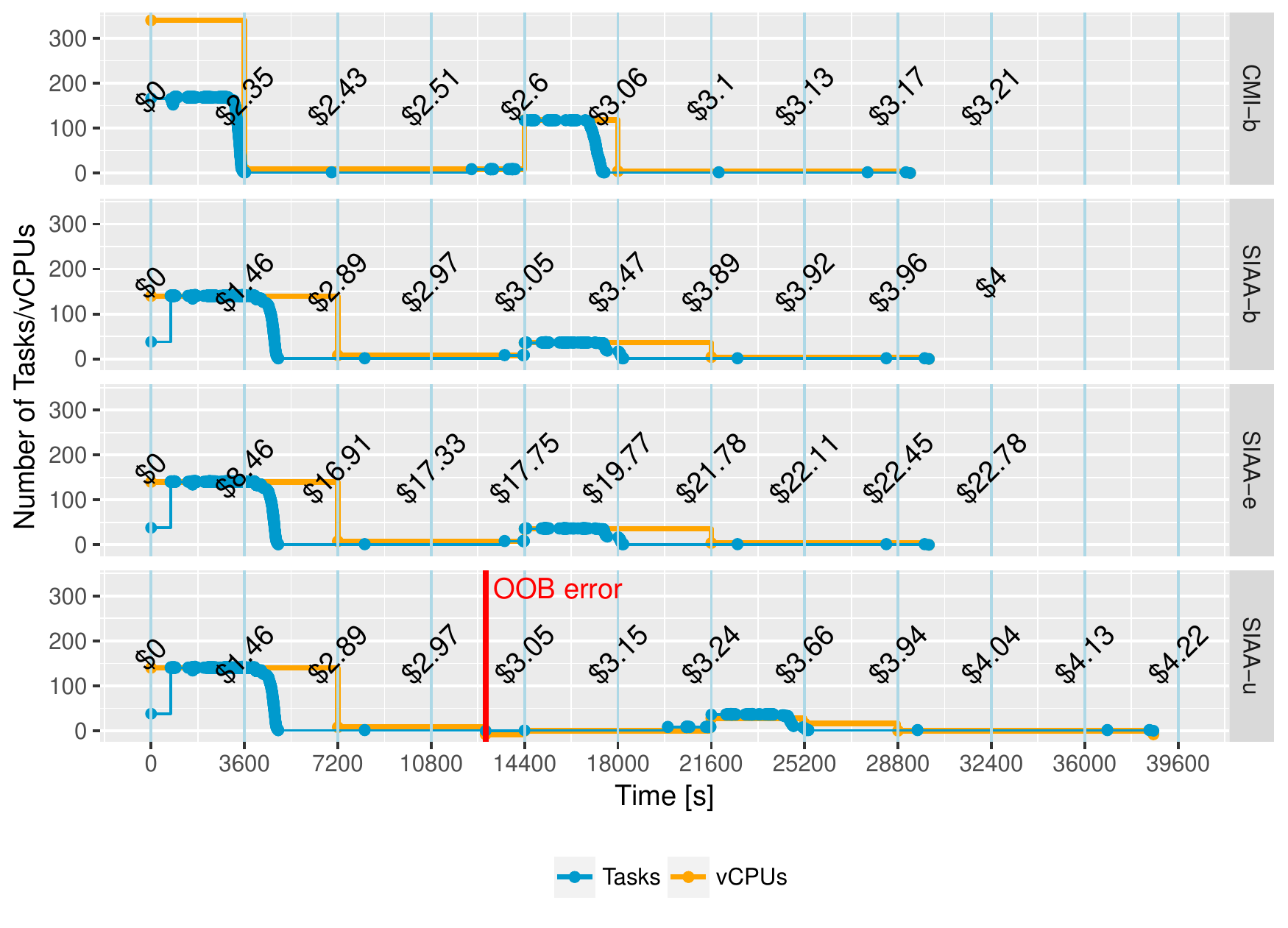}
\par\end{centering}
\caption{\label{fig:Autoscaling-analysis}Analysis of the autoscaling process
for Montage for CMI-b, the best performing configuration of SIAA (SIAA-b),
a configuration of SIAA that produces a high cost of execution (SIAA-e),
and a configuration of SIAA that produces an OOB error affecting the
execution time (SIAA-u). The vertical red line indicates the occurrence
of an OOB error during the execution of SIAA-u.}
\end{figure}
\end{enumerate}

\subsection{\label{subsec:Statistical-significance}Statistical Significance
of Results}

In this subsection we show the statistical significance of the obtained
results. Results are statistically significant when there is a possibility
that a relationship between two or more variables is caused by something
other than chance. The statistical hypothesis testing is used to determine
if the result of a dataset is statistically significant. This test
provides a p-value, representing the probability that a random chance
could explain the result. This kind of analysis is fundamental to
support the validity of the performance differences found between
the analyzed autoscaling strategies.

Table~\ref{tab:L2-distance} shows the mean, median, standard deviation~(SD),
min. and max. for the $L_{2}$-norm metric per workflow. Centrality
metrics (mean and median) and max. values of the $L_{2}$-norm favors
CMI in all cases, in terms of SD and min. value, CMI has the smallest
values in most cases.

\begin{table}[h]
\caption{\label{tab:L2-distance}Detailed $L_{2}$-norm results. Each row presents
the mean values per strategy considering all the studied applications.
Best mean and median values are highlighted. Last column describes
the result of the normality test Shapiro-Wilk with a confidence level
of $\alpha=0.001$.}

\centering{}%
\begin{tabular*}{1\textwidth}{@{\extracolsep{\fill}}lccccccc}
\toprule 
Workflow & Strategy & Mean & Median & SD & Min. & Max. & Normal\tabularnewline
\midrule
\midrule 
CyberShake & CMI & \textbf{0.22} & \textbf{0.17} & 0.12 & 0.10 & 0.69 & no\tabularnewline
 & SIAA & 0.71 & 0.83 & 0.26 & 0.08 & 1.01 & no\tabularnewline
\midrule 
Inspiral & CMI & \textbf{0.14} & \textbf{0.09} & 0.16 & 0.01 & 0.64 & no\tabularnewline
 & SIAA & 0.72 & 0.85 & 0.25 & 0.02 & 1.15 & no\tabularnewline
\midrule 
Montage & CMI & \textbf{0.10} & \textbf{0.07} & 0.07 & 0.02 & 0.25 & no\tabularnewline
 & SIAA & 0.80 & 0.92 & 0.20 & 0.18 & 1.07 & no\tabularnewline
\midrule 
Pan-STARRS & CMI & \textbf{0.39} & \textbf{0.39} & 0.23 & 0.01 & 1.01 & yes\tabularnewline
 & SIAA & 0.63 & 0.66 & 0.14 & 0.27 & 1.03 & no\tabularnewline
\bottomrule
\end{tabular*}
\end{table}

To evaluate the significance of the improvements achieved by CMI we
applied the Mann\textendash Whitney \emph{U} test~\citet{Mann1947},
a non-parametric method that evaluates if two data samples come from
the same population\footnote{As the $L_{2}$-metric results are non-normally distributed (See Table~\ref{tab:L2-distance})
a non-parametric method is required to test the significance of the
differences.}. To ensure the validity of our claims we selected a strong confidence
level of $\alpha=0.001$. We compared CMI versus each of the 55 configurations
of SIAA. Table~\ref{tab:Summary-of-statistical-tests} summarizes
the number of times CMI outperforms SIAA in terms of the $L_{2}$-norm,
the number of times that improvement is significant and the range
of percentual improvements for each application. 

For each workflow, we carried out 11 tests corresponding to the comparison
of CMI with each variant of SIAA for the different spot ratio (SR).
In the table, the ``Better'' column describes the number of cases
for which CMI presented significant improvements over the different
SIAA configurations in terms of the $L_{2}$-norm. The ``Non signif.''
column presents the number of cases in which there were not significant
differences between CMI and SIAA in terms of the $L_{2}$-norm. The
columns ``Min impr.'' and ``Max impr.'' show the range of percentual
improvements that are significant according to the test. It is important
to mention that in none of the cases SIAA presented statistically
significant improvements over CMI.

\begin{table}[h]
\caption{\label{tab:Summary-of-statistical-tests}Summary of statistical test
results (Mann-Whitney U test with a confidence level of $\alpha=0.001$).
The amount of tests for which CMI present significant improvements
over SIAA, are highlighted. }

\centering{}%
\begin{tabular*}{1\textwidth}{@{\extracolsep{\fill}}lccccc}
\toprule 
Workflow & Total no. of tests & Better & Non signif. & Min impr. & Max impr.\tabularnewline
\midrule
\midrule 
\Cyb{} & 11 & \textbf{10} & 1 & 14.09\% & 70.24\%\tabularnewline
\Lig{} & 11 & \textbf{11} & 0 & 31.54\% & 80.04\%\tabularnewline
\Mon{} & 11 & \textbf{11} & 0 & 34.68\% & 85.77\%\tabularnewline
\Pan{} & 11 & \textbf{9} & 2 & 10.16\% & 38.75\%\tabularnewline
\midrule 
All & 44 & \textbf{41} & 3 & 10.16\% & 85.77\%\tabularnewline
\bottomrule
\end{tabular*}
\end{table}

This is expected as the Pan-STARRS workflow comprises very long tasks
(almost 7 days in the worst case), not captured by our definition
of the probabilities computation since it considers one-hour time
windows. Therefore our model greatly underestimates error probabilities
for such tasks impacting mainly on makespan as we saw in Table~\ref{tab:summary}.
We believe that simply computing probabilities in a more fine-grained
way might allow us to obtain better autoscaling solutions allowing
CMI to outperform SIAA with wider margins.

\section{\label{sec:Related-work}Related Work}

In the literature there are many approaches addressing the efficient
management of scientific applications on the cloud~\citet{Netto2014},
however, to the best of our knowledge, there are no efforts covering
\emph{autoscaling} strategies based on an online multi-objective meta-heuristic
for executing \emph{workflow} applications, which additionally consider
spot instances and minimize both the makespan, cost and probability
of OOB errors. Among these works, we can mention our own past work
in~\citet{Monge2017} where we proposed an autoscaler called SIAA
for workflow applications which considered spot instances. The aim
of~\citet{Monge2017} was to minimize the makespan subject to budget
constraints (i.e. limiting the maximum cost). In such work, to cope
with spot instances unreliability, tasks were heuristically scheduled
with the aim of mitigating the negative effects of out-of-bid errors
on the workflow makespan. The main difference with respect to the
present paper is that in~\citet{Monge2017} the monetary cost was
not considered. Then, in~\citet{Li2015}, the authors have proposed
a cost-efficient based scheduling algorithm which allows to lease
instances from clouds for executing scientific workflows while meeting
the required deadlines of tasks. The tasks are scheduled according
to the spot instance pricing. On the other hand, the work in~\citet{Lu2013}
is focused on running large-scale computational applications on clouds,
especially for on-demand instances and spot instances offered by Amazon
EC2. In~\citet{Lu2013}, after analyzing the characteristic of the
spot price and the effect of spot instances disturbance, the authors
proposed a dynamic approach for running the applications with the
aim of reducing cost, increasing the reliability and reducing the
complexity of fault tolerance without affecting the overall performance
and scalability. However, the main difference between the works~\citet{Li2015,Lu2013}
and ours is that we focus on a budget-constrained autoscaling problem
while the efforts mentioned focus on solving scheduling problems subject
to task deadline constraints, thus they are useful in different scenarios.
Another important distinction is that we are also focused in reducing
the probability of failures \textendash to avoid heavily affecting
workflow application results\textendash{} and without relying on spot
price predictors.

There are also some works particularly addressing \emph{workflow autoscaling}~\citet{DeConinck2016,Ding2017,Mao2011,Mao2013}
with deadline or budget constraints. The problem was first addressed
by~\citet{Mao2011} where the authors proposed an autoscaling strategy
for the efficient execution of multiple workflow applications subject
to deadline constraints. The goal of~\citet{Mao2011} was to ensure
that all tasks finish before their respective deadlines by using the
cheapest resources whenever possible. Later, the same authors moved
to the problem of workflow autoscaling but considering budget constraints~\citet{Mao2013}.
Another work is the one presented in~\citet{DeConinck2016}, where
the main characteristics of the tasks in the workflow structure are
learned over time, i.e., the autoscaler dynamically adapts the number
of allocated resources in order to meet the deadlines of all tasks
without knowing the workflow structure itself and without any information
of the execution time. The goal of this work was to minimize the makespan.
These strategies were proposed for simultaneously addressing the problem
of scaling down/up the virtual cloud infrastructure and scheduling
workflow tasks in heterogeneous cloud infrastructures but they did
not consider spot instances, which is one of the main differences
with respect to our work.

It is worth noting that, from the related works found, most of them
have been proposed for workflows considering task deadlines or budget
constraint but without considering spot instances. Besides, another
crucial distinction is that in none of the surveyed works the authors
have considered to reduce the impact of OOB errors as part of the
optimization process when using spot instances, and only in three
works~\citet{DeConinck2016,Qu2016,Poola2014a} the authors have proposed
to minimize the makespan, rendering difficult their applicability
to execute scientific workflows in clouds infrastructures to support
rapid domain-specific ``in vitro'' experimentation. Concretely,
in the present paper the objectives are to minimize the makespan,
monetary cost and the failure probability when different types of
spot instances and bid prices are considered.

\section{\label{sec:Conclusions}Conclusions and Future Work}

Due to the complexity of scientific and engineering processes, scientific
and engineering workflows have become increasingly challenging from
a computational standpoint. Then, such workflows are often required
to be executed in a distributed high-performance computing environment
such as a cloud. A cloud combines the customization of instances with
scalability and resource sharing capabilities. Moreover, clouds allow
users the acquisition of instances under a pay-per-use scheme and
where the prices differ according to the type of instance acquired
and the pricing model of the cloud provider. Concretely, there are
two pricing models through which instances can be acquired, i.e. on-demand
and spot instances.

In this work we have proposed a new autoscaler for scientific and
engineering workflows denominated CMI. This autoscaler exploits the
elasticity potential of cloud infrastructures by approximating the
right amount of on-demand and spot instances. To cope with the dynamically
changing computing demands of workflows during execution, CMI periodically
adjusts the number of instances of each type and determines the adequate
bid prices to acquire spot instances aiming at minimizing makespan,
monetary cost and the potential impact of task failures. These optimization
problems are solved by using the popular meta\textendash heuristic
NSGA-II. A sensitivity analysis of NSGA-II was performed to determine
the best hyper-parameter set for each workflow application, together
with detailed guidelines regarding how this analysis could be conducted
for other workflows.

Simulation experiments on 4 large\textendash scale scientific and
engineering workflows indicate that CMI outperforms a closely-\-related
state-of-the-art heuristic autoscaling strategy called SIAA in terms
of the $L_{2}$-norm of makespan and cost. In this work, in order
to perform a rigorous comparative study, we set up SIAA with 55 different
configurations involving different combinations between on\textendash demand
and spot instances and different bidding strategies that trade\textendash off
cost and reliability. Statistical tests show that CMI outperforms
SIAA in 41 of 44 cases, with statistically significant improvements
regarding the $L_{2}$-norm metric. In the remaining 3 cases differences
are not statistically significant. We believe these results are highly
encouraging and of practical relevance.

This work can be extended by deepening the study of spot prices and
failure probabilities due to OOB errors. Obtaining a more informative
model of such errors that takes into account the expected duration
of tasks will potentially help in the reduction of failures as bid
prices could be better fitted to the expected duration of tasks. We
expect that such a model could lead to improving the overall performance
of CMI.

Another aspect that deserves special attention is to explore the ideas
exposed in this article in the context of other evolutionary meta\textendash heuristic
algorithms such as NSGA-III, $\epsilon$\textendash MOEA, MOEA/D,
which are more suitable for solving many\textendash objective problems
than NSGA-II. Performing an extensive comparison of algorithms robustness
under different settings and varieties of workflows will provide a
deeper understanding of the characteristics of the problem and their
solutions.

Finally, we are interested on extending the mathematical model of
the autoscaling problem to consider data storage cost and storage
access times. Designing a data\textendash aware formulation for the
autoscaling problem is fundamental for running today's data\textendash intensive
scientific workflows, and in this way not only supporting CPU-intensive
workflows. In such context the occurrence of OOB errors might seriously
handicap makespan derived from to the need of re-transmissions inside
the cloud of huge amounts of data but also might critically increase
execution costs due to extra use of storage/network resources, which
must be rented in pay-per-use clouds.

\section*{Reproducibility}

Workflow descriptions are available on-line through the Pegasus WorkflowGenerator\footnote{WorkflowGenerator: \url{http://confluence.pegasus.isi.edu/display/pegasus/WorkflowGenerator}}.
Spot prices data used in this experiment are publicly available~\citet{SpotsData2016}.
Experiments performed in this study were run on an Intel Core~i5
computer running Ubuntu Desktop~16.0 and Java version~\JavaVersion{}.
Implementation of the NSGA-II algorithm and the tools for the sensitivity
analysis were provided by the MOEA Framework\footnote{MOEA Framework: \url{http://moeaframework.org/}}
version~\MoeaVersion{}. Simulations were performed using the CloudSim\footnote{CloudSim: \url{http://www.cloudbus.org/cloudsim/}}
simulator version~\CloudSimVersion{}. We used implementation of
the Mann\textendash Whitney U test incorporated in Apache Commons
Math\footnote{Commons-Math: \url{http://commons.apache.org/proper/commons-math/}}
version~\CommonsMathVersion{}~ for evaluating the statistical significance
of our results. Visualizations were generated using Tableau~Public\footnote{Tableau Public: \url{http://public.tableau.com}}
version~\TableauVersion{} and ggplot2\footnote{ggplot2: \url{http://ggplot2.tidyverse.org}}
version~\ggplotVersion{}.

\section*{Acknowledgments}

This research is supported by the ANPCyT projects No. PICT-2012-2731,
PICT-2014-1430 and PICT-2015-1435, and by the SIIP-UNCuyo project
No.~M041. This research has been partially funded by the Spanish
Ministry of Science and Innovation and FEDER under contract TIN2017-88213-R
and the network of smart cities CI-RTI (TIN2016-81766-REDT).

\bibliographystyle{unsrt}
\bibliography{biblio}

\end{document}